\pdfoutput=1

\documentclass[11pt]{article}

\usepackage[]{acl}

\usepackage{times}
\usepackage{amsmath}
\usepackage{cleveref}
\crefname{section}{§}{§§}
\Crefname{section}{§}{§§}
\usepackage{latexsym}
\usepackage{framed}
\usepackage{color}
\usepackage{xcolor}
\usepackage{tcolorbox}
\tcbuselibrary{most}
\tcbset{on line, 
        boxsep=4pt,left=0pt,right=0pt,top=0pt,bottom=0pt,
        colframe=white, 
        highlight math style={enhanced} %
        }
\definecolor{shadecolor}{rgb}{0.92,0.92,0.92}
\usepackage{multirow}
\usepackage{booktabs}

\usepackage{amsfonts}
\usepackage{amssymb}
\usepackage{graphicx}
\usepackage{caption}
\usepackage{subcaption}
\usepackage{url}
\usepackage[T1]{fontenc}

\usepackage{microtype}

\title{Prototypical Verbalizer for Prompt-based Few-shot Tuning}

\author{Ganqu Cui$^{1,2}$, Shengding Hu$^{1,2}$, Ning Ding$^{1,2}$, Longtao Huang$^{6}$,  Zhiyuan Liu$^{1,2,3,4,5}$\thanks{\quad Corresponding author: Z.Liu (liuzy@tsinghua.edu.cn)}\\
$^{1}$Dept. of Comp. Sci. \& Tech., Institute for AI, Tsinghua University, Beijing, China\\
$^{2}$Beijing National Research Center for Information Science and Technology\\
$^{3}$Institute Guo Qiang, Tsinghua University, Beijing, China\\
$^{4}$International Innovation Center of Tsinghua University, Shanghai, China\\
$^{5}$Beijing Academy of Artificial Intelligence
$^{6}$Alibaba Group\\
{\tt cgq19@mails.tsinghua.edu.cn}
}

\begin{document}
\maketitle
\begin{abstract}
Prompt-based tuning for pre-trained language models (PLMs) has shown its effectiveness in few-shot learning. Typically, prompt-based tuning wraps the input text into a cloze question. 
To make predictions, the model maps the output words to labels via a \textit{verbalizer}, which is either manually designed or automatically built. However, manual verbalizers heavily depend on domain-specific prior knowledge and human efforts, while finding appropriate label words automatically still remains challenging.
In this work, we propose the prototypical verbalizer (ProtoVerb) which is built directly from training data. Specifically, ProtoVerb
learns prototype vectors as verbalizers by contrastive learning. In this way, the prototypes summarize training instances and are able to enclose rich class-level semantics. 
We conduct experiments on both topic classification and entity typing tasks, and the results demonstrate that ProtoVerb significantly outperforms current automatic verbalizers, especially when training data is extremely scarce. More surprisingly, ProtoVerb consistently boosts prompt-based tuning even on untuned PLMs, indicating an elegant non-tuning way to utilize PLMs. Our codes are avaliable at \url{https://github.com/thunlp/OpenPrompt}.
\end{abstract}

\section{Introduction}
\label{sec:intro}

\begin{figure}[h]
    \centering
    \includegraphics[width=\linewidth]{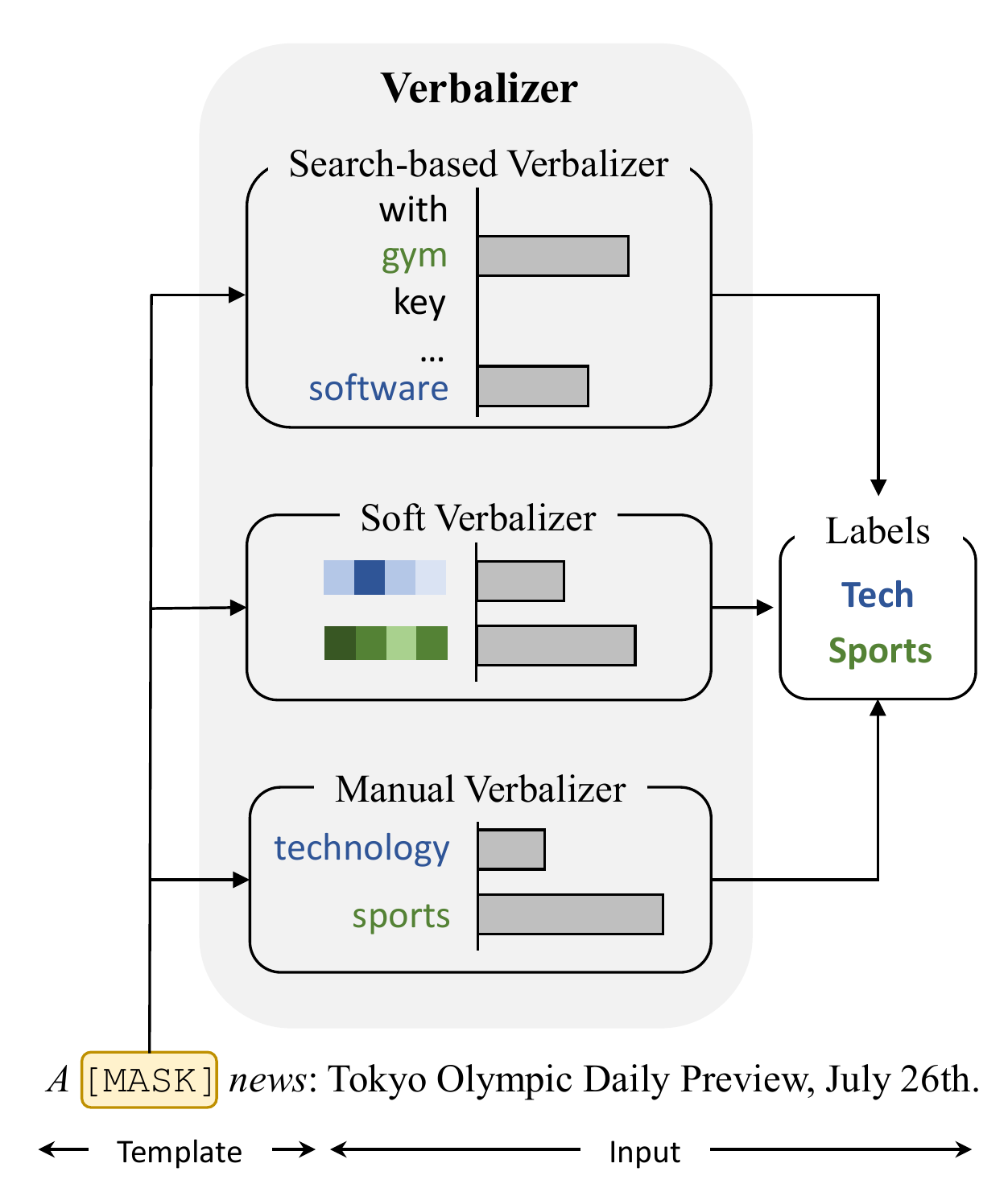}
    \caption{Illustration of three verbalizer construction methods.}
    \label{fig:verbalizer}
\end{figure}

The massive-scale pre-trained language models (PLMs)~\cite{han2021pre}
have been proven to be backbones for solving a variety of NLP tasks~\cite{kowsari2019text, rajpurkar2016squad}. 
To further adapt these PLMs to downstream tasks such as classification, traditional approaches fine-tune the language models through an extra classifier~\cite{howard2018universal}. However, when task-specific data is limited~\cite{bragg2021flex},  training the extra classifier effectively is challenging due to the gap between pre-training tasks (e.g., masked language modeling) and fine-tuning tasks (e.g., classification and regression). This gap impedes the fast adaptation of PLMs to downstream tasks. 

Recently, prompt-based tuning~\cite{schick21pet, liu2021gpt} has risen to be a powerful way for few-shot learning by bridging the gap between the pre-training stage and downstream task stage. In prompt-based tuning, the input texts are wrapped with task-specific templates to re-formalize the original task as a cloze-style task. For example, in topic classification task, we can use template ``\texttt{<text>} This topic is about \texttt{[MASK]}'', where \texttt{<text>} is the placeholder for input sentences. The PLMs are asked to infer the words to fill in \texttt{[MASK]} and the words are further mapped to corresponding labels through a \textit{verbalizer} (e.g. ``sports'' for label ``Sports''). Verbalizers are of great importance in prompt-based tuning~\citep{gao20making} since they are the bridges between model outputs and the final predictions. How to build effective verbalizers for prompt-based tuning---especially for many-class classification, is a critical issue in prompt-based tuning.

Typically, most current works adopt three kinds of verbalizers: manual verbalizers, search-based verbalizers, and soft verbalizers. We show them by an example in Figure~\ref{fig:verbalizer}. Human-designed manual verbalizers pick some label words (e.g. label names) to depict classes. These verbalizers are powerful across multiple tasks~\citep{schick21pet}. Despite their success, a major drawback roots in the strong assumption that we own precise understandings of downstream tasks and are able to sum up each class with several words.
Without task-specific prior knowledge, selecting appropriate label words is non-trivial. Further, they also need intensive human labors when facing many classes.
To mitigate these issues, search-based verbalizers aim at finding suitable label words from vocabulary with algorithms ~\citep{schick20petal, shin2020autoprompt, gao20making} and soft verbalizers use trainable tokens which are optimized during tuning~\citep{hambardzumyan21warp, zhang2021differentiable}. However, it is challenging to search or optimize adequately in a large vocabulary or embedding space under a low-data regime, making automatic verbalizers suboptimal compared with manual ones.


Intuitively, class proxies in verbalizers should encapsulate class-level semantic features, which are expressed implicitly by instances. To obtain these semantic representatives with few data, one promising approach is computing central points of class instances, namely \textit{prototypes}, as approximation.
To this end, we manage to estimate prototype vectors for each class to serve as verbalizer.
Summarized from instances, prototypes are supposed to establish concepts similar with human-designed labels.

In this work, we introduce prototypes into this problem and propose prototypical verbalizer (ProtoVerb), which learns class prototypes from training data to build verbalizers automatically. For prototype learning, inspired by the idea of PCL~\citep{li2021proto}, ProtoVerb trains the prototype vectors by contrastive learning with the InfoNCE estimator~\citep{oord2018representation}. Specifically, our optimization objective includes two components: The first part is an instance-instance loss to cluster intra-class instances and separate inter-class instances; The second part is an instance-prototype loss which enforces the prototypes to be center points of classes. 
Compared with other verbalizer construction methods, ProtoVerb learns continuous vectors straight from training instances efficiently, which makes it a plug-in-and-play algorithm with high flexibility.

To verify the effectiveness of ProtoVerb, we conduct extensive experiments on topic classification and entity typing tasks. We study two different settings where ProtoVerb can work: (1) When manual verbalizers are available, ProtoVerb can play as an extra verbalizer in the inference stage. Results show that ProtoVerb consistently improves the classification performance with low cost, and even untuned PLMs benefit largely.
(2) Consider a realistic setting where only a limited number of samples are provided with no manual verbalizers, ProtoVerb also produces verbalizers of high quality. Experimental results demonstrate that ProtoVerb significantly outperforms existing search-based and soft verbalizers.

\section{Related Work}
\subsection{Prompt-based Tuning}
Despite the success of PLMs~\citep{Devlin19bert, liu2019roberta, raffel2019exploring} in massive NLP tasks, few-shot fine-tuning of PLMs was suboptimal due to the gap between pre-training and downstream tasks. Inspired by the ``in context learning'' proposed by GPT-3~\citep{Brown20language}, stimulating model knowledge with a few \textit{prompts} has recently received much attention. A series of prompt-based work on knowledge probing~\citep{trinh2018simple, Petroni19lama, Davison19common}, text classification~\citep{schick21pet, gao20making}, relation extraction~\citep{han2021ptr}, and entity typing~\citep{ding2021prompt} emerge and achieve impressive progress. 
Typically, a piece of prompt contains a template and a verbalizer. Early prompts employ human-picked prompts which demand human knowledge and manual efforts. To alleviate this issue, later works explore automatic designing and optimizing prompts~\citep{liu2021gpt, gao20making, zhang2021differentiable}. Recently research works further propose continuous prompts to replace the discrete phrases~\citep{lester2021power, LiL2021prefix}. However, the designation of verbalizers, an important part of prompts, is less explored.
In this work, we investigate the automatic verbalizer construction in prompt-based tuning.

\subsection{Verbalizer Design}
Verbalizers bridge between model outputs and labels and make great impact on prompt-based tuning~\citep{gao20making}. With task-specific knowledge, human-picked words are widely used and proved effective~\citep{schick21pet}. The major drawback of manual verbalizers is the assumption that we possess sufficient knowledge of downstream tasks, which is not always satisfied.
To avoid intensive human labor and expert knowledge dependency in manual verbalizers, some works explore search-based verbalizers~\citep{schick20petal, gao20making, shin2020autoprompt} that identify label words automatically with training data. However, with a large vocabulary and few examples, it is non-trivial to find suitable words.
Another line of researches focuses on soft verbalizers~\citep{hambardzumyan21warp, zhang2021differentiable}, which insert continuous embeddings as soft labels. The label embeddings are optimized along with model tuning. Similarly, soft verbalizers require abundant data for sufficient optimization, which can not be satisfied with the few-shot setting. In contrast, our approach learns prototype vectors from scratch, hence is more effective for few-shot tuning.

\subsection{Prototype-based Few-shot Learning}

In few-shot learning, prototype-based metric-learning methods have been promising approaches for their simplicity and effectiveness. 
Prototypical Networks (ProtoNet)~\citep{Snell17proto}  is the pioneering work that introduces prototypes into deep learning. Specifically, ProtoNet calculates prototype vectors by taking the average of instance vectors and makes predictions by metric-based comparisons between prototypes and query instances. A set of following works concentrates on the advancement of prototype estimation~\citep{li2021proto, Gao19hybrid, Ding21prototypical}. Among them, PCL~\citep{li2021proto} achieves remarkable results on self-supervised few-shot learning by using prototypes as latent variables and inspires us in designing training objectives.
The success of prototype-based models indicates that prototypes, which are representative embeddings of instances from the same classes, encapsulate some class-level semantic features. Inspired by the intrinsic similarity of prototypes and verbalizers, we find it natural and elegant to introduce prototypes into verbalizer construction for prompt-based tuning.

\section{Background}
Given a pre-trained language model $\mathcal{M}$, our goal is to tune it for specific downstream tasks. Take $N$ way $K$ shot few-shot text classification as an example, the support set for class $n$  $\mathcal{D}_n=\{x^n_1,\cdots, x^n_K\}$ contains $K$ sentences. We aim to predict the label $y\in\mathcal{Y}$ for each sentence, where $\mathcal{Y}$ is the label set with $N$ distinct classes.

\subsection{Fine-tuning}
For a sentence concatenated with special tokens $x=\{\texttt{[CLS]}, t_1, \cdots, t_T, \texttt{[SEP]}\}$, language model $\mathcal{M}$ encodes it into hidden representations $\{\mathbf{h}_{\texttt{[CLS]}}, \mathbf{h}_1, \cdots, \mathbf{h}_T, \mathbf{h}_{\texttt{[SEP]}}\}$.
Conventional fine-tuning trains an extra classifier $F$ over the \texttt{[CLS]} embedding $\mathbf{h}_{\texttt{[CLS]}}$ and output the probability distribution on label set $\mathcal{Y}$.
\begin{equation}
    P(\cdot | x) = \text{Softmax}(F(\mathbf{h}_{\texttt{[CLS]}})).
\end{equation}

The classifier and PLM are tuned by maximizing $\frac1N\sum_{i=1}^N \log P(y_i|x_i)$, where $y_i$ is the label of $x_i$.

\begin{figure*}
    \centering
    \includegraphics[width=\linewidth]{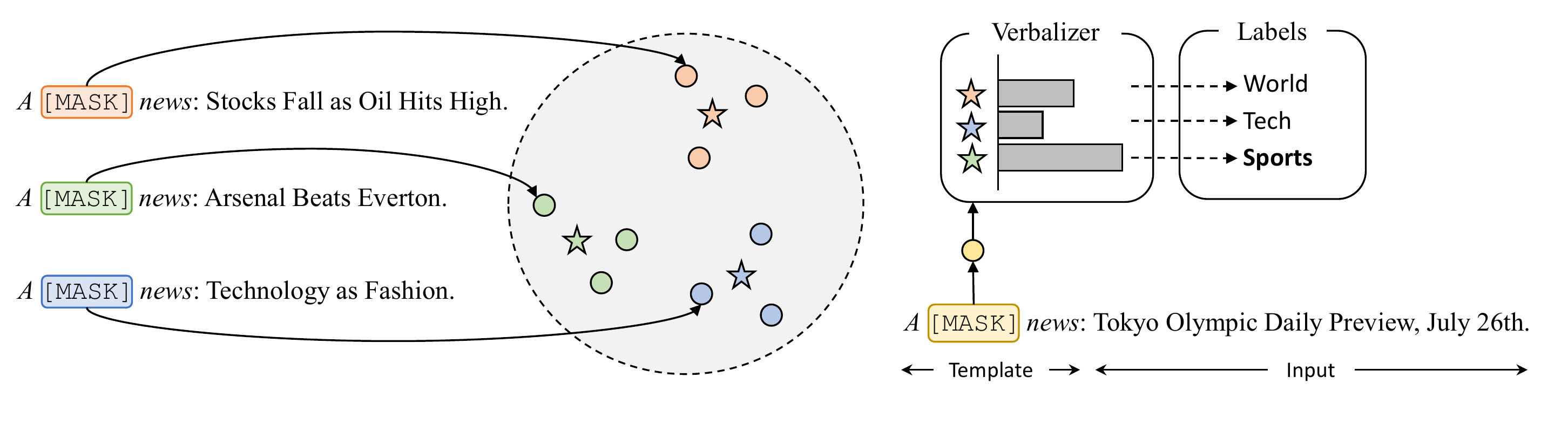}
    \caption{Illustration of ProtoVerb. Left: We project the hidden states of \texttt{[MASK]} tokens to the embedding space and learn prototypes. Right: The learned prototypes constitute the verbalizer and map the PLM outputs to corresponding labels.}
    \label{fig:model}
\end{figure*}

\subsection{Prompt-based Tuning}
The vanilla prompt-based tuning converts the downstream task to a cloze-style mask language modeling problem. For example, to formulate the text classification task, we can modify the original input $x$ with a template $\mathcal{T}(\cdot) = \text{A \texttt{[MASK]} news: }$ to get the prompt input $\mathcal{T}(x) = \text{A \texttt{[MASK]} news: }x$. With  $\mathcal{T}(x)$, $\mathcal{M}$ produces the hidden vector at the \texttt{[MASK]} position $\mathbf{h}_\texttt{[MASK]}$. To calculate the probability distribution over the label set, a manual verbalizer stores a set of label words $\mathcal{V}$ and the score for label $y$ is 
\begin{equation}
    P_{\mathcal{M}}(y|x)=g(P_{\mathcal{M}}(\texttt{[MASK]}=v|\mathcal{T}(x))|v\in \mathcal{V}_y),
\end{equation}
where $\mathcal{V}_y$ is the label words of $y$ and $g(\cdot)$ is to aggregate multiple scores.

\section{Prototypical Verbalizer}

In previous sections, we introduce the general pipeline of prompt-based tuning. As manually defining or automatically searching for appropriate verbalizers can be challenging, here we propose to learn prototypes directly from training instances. Inspired by PCL~\citep{li2021proto}, the prototypes are trained with contrastive learning. As shown in Figure~\ref{fig:model}, we first get the hidden states of \texttt{[MASK]} tokens to represent instances, then project them to another embedding space for prototype learning. The prototypes are used as verbalizers for prediction.
Next, we will introduce the learning and inference stages of ProtoVerb in detail.

\subsection{Instance Representation and Similarity Function}

Given a piece of training text $x$ wrapped with a template, we take the last layer's hidden state of the \texttt{[MASK]} token $\mathbf{h}_\texttt{[MASK]}$ as the initial representation of the text. With an encoder $E_{\phi}(\cdot)$ parameterized by $\phi$, the instance representation of $x$ is
\begin{equation}
    \mathbf{v} = E_{\phi}(x)=\mathbf{W}\mathbf{h}_\texttt{[MASK]}.
\end{equation}

In practice, we simply adopt a linear encoder with weight $\mathbf{W}$. To measure the similarity between instances, we adopt cosine similarity function $S(\cdot)$, where 
\begin{equation}
    S(\mathbf{v}_i, \mathbf{v}_j) = \frac{\mathbf{v}_i}{||\mathbf{v}_i||} \cdot \frac{\mathbf{v}_j}{||\mathbf{v}_j||}.
\end{equation}

\subsection{Loss Function}
With the instance representation and similarity function, we discuss how to define our training objective. Denote $\mathcal{C}=\{\mathbf{c}_1, \cdots, \mathbf{c}_N\}$ as the set of prototype vectors. Intuitively, there are two goals we need to achieve by optimization: (1) For instance-instance pairs, intra-class pairs should get higher similarity scores than inter-class pairs. (2) For instance-prototype pairs, the similarity scores between prototype $\mathbf{c}_n$ and instances of class $n$ should be higher than $\mathbf{c}_n$ and other instances. To realize these two goals, we define the objective function based on the InfoNCE estimator~\citep{oord2018representation}, which is widely adopted in contrastive learning. 

For the instance-instance objective, we minimize the following loss function
\begin{equation}
    \mathcal{L}_{\text{ins}} = \frac{-1}{N^2K^2} \sum_{n} \sum_{i,j} \log\frac{\exp{S(\mathbf{v}^n_i, \mathbf{v}^n_j)}}{\sum_{n\prime,j\prime}\exp{S(\mathbf{v}^n_i, \mathbf{v}^{n\prime}_{j\prime})}},
\end{equation}
where $(\mathbf{v}^n_i, \mathbf{v}^n_j)$ are instance pairs of the same class. This loss function maximizes  intra-class similarity and minimizes inter-class similarity between instances.

Similarly, the instance-prototype loss function is defined as 
\begin{equation}
    \mathcal{L}_{\text{proto}} = \frac{-1}{N^2K}\sum_{i,n} \log\frac{\exp{S(\mathbf{v}^n_i, \mathbf{c}_n)}}{\sum_{n\prime}\exp{S(\mathbf{v}^n_i, \mathbf{c}_{n\prime})}},
\end{equation}
and $\mathbf{v}^n_i$ is of class $n$. This objective forces each prototype to lie at the center point of its instances. 

Overall, combining the instance-instance loss and instance-prototype loss, our final training objective is 
\begin{equation}
    \mathcal{L} = \mathcal{L}_{\text{ins}} + \mathcal{L}_{\text{proto}}.
\end{equation}

\subsection{Inference}
During inference, following the same metric, we calculate the similarity scores of query and prototypes. The probability score for class $k$ is 
\begin{equation}
    P_{\mathcal{M}}(y_k|x)=\frac{\exp{S(\mathbf{v}, \mathbf{c}_k)}}{\sum_{k\prime}\exp{S(\mathbf{v}, \mathbf{c}_{k\prime})}}.
\end{equation}

Then we make prediction by $\arg \max$ function
\begin{equation}
    \widetilde{y} = \mathop{\arg \max}_{k} P_{\mathcal{M}}(y_k|x).
\end{equation}

When there are other verbalizers (e.g. manual verbalizers), we first process the logits from different verbalizers with a standard scaler (minus mean then divide by standard deviation). Then we take the mean value of the scores to get the final score.

\section{Experiments}
We conduct extensive few-shot learning experiments to illustrate the effectiveness of ProtoVerb. In this section, we first introduce the experimental settings in use. Then we present and discuss the experiment results.


\subsection{Datasets and Templates}
Verbalizers in many-class classification tasks are difficult to get precise definitions. Hence
we adopt three topic classification datasets: AG's News, Yahoo~\citep{zhang2015character}, and DBPedia~\citep{lehmann2015dbpedia} and one entity typing dataset: FewNERD~\citep{ding21fewnerd} as benchmarks, and their statistics are summarized in Table~\ref{tab:dataset}. 

To focus on the verbalizer and alleviate the influence of templates, we adopt multiple fixed manual templates. 
For topic classification, following~\citep{hu2021knowledgeable}, we use four templates on each dataset. For entity typing, we use three templates from~\citep{ding2021prompt}. Details about the templates can be found in Appendix~\ref{sec:appendix}.

\begin{table}[ht]
    \centering
    \begin{tabular}{l|c|c|c}
    \toprule
        Dataset & Task & \#Class & \#Test \\
        \midrule
        AG's News & TC & 4 & 7,600 \\
        DBPedia & TC & 14 & 70,000\\
        Yahoo & TC & 10 & 60,000\\
        FewNERD & ET & 66 & 96,901\\
        \bottomrule
    \end{tabular}
    \caption{Dataset statistics. TC is for topic classification and ET is for entity typing.}
    \label{tab:dataset}
\end{table}

\subsection{Experimental Settings}
Under the few-shot setting, we randomly sample $k=1,2,4,8,16$ instances in each class from the training set and test the model on the entire test set. As for the evaluation metric, we use accuracy in all experiments.
For the different usages of ProtoVerb, we consider two specific settings:

(1) ProtoVerb as a single verbalizer (\cref{sec:single}). When manual verbalizers are not available, we can tune the model with ProtoVerb. Under this setting, we want to evaluate the performance of ProtoVerb compared with other automatic verbalizer construction methods.

(2) ProtoVerb as an extra verbalizer (\cref{sec:extra}). Naturally, we suppose that there exists a manual verbalizer and we append ProtoVerb to strengthen the performance. Under this setting, ProtoVerb is a plug-in-and-play component and does not participate in the tuning process. We compare ProtoVerb with manual verbalizers and other verbalizer ensembles.

\subsection{Implementation Details}
All our models and baselines are implemented with PyTorch~\citep{paszke19pytorch} framework, Huggingface transformers~\citep{wolf2020transformers}, and OpenPrompt toolkit~\citep{ding2021openprompt}. We optimize PLMs with AdamW optimizer~\citep{loshchilov19decoupled}. For prototype learning, we set the prototype dimension to 128 and optimize the loss function with Adam optimizer~\citep{Kingma14adam}.
For topic classification, we use RoBERTa-large~\citep{liu2019roberta} as our PLM backbone and tune the model for 5 epochs. The batchsize is 2 and the learning rate is 3e-5.
For entity typing, we tune a BERT-base~\cite{Devlin19bert} model for 30 epochs and set the batchsize to 16. The learning rate here is 5e-5.

\subsection{Baselines}
The vanilla prompt-based tuning method fuses the input text with a task-specific template and maps the model outputs to labels through a verbalizer. For fair comparisons, all our baselines and proposed models are built on this pipeline and they merely differ from the verbalizers in use.

\textbf{Manual verbalizers} (ManualVerb) are defined by human with domain knowledge. Here we simply employ the verbalizers provided by OpenPrompt~\citep{ding2021openprompt}.

\textbf{Search-based verbalizers} (SearchVerb) search for suitable words from vocabulary automatically. We adopt the implementation in PETAL~\citep{schick20petal}, which finds the words that maximize the likelihood of the training data. To combine SearchVerb with ManualVerb, we merge their verbalizer words together.

\textbf{Soft verbalizers} (SoftVerb) introduce trainable tokens as verbalizers in prompt-based tuning. We follow the approach in WARP~\citep{hambardzumyan21warp} that applies soft tokens as a linear decoding layer, and the token embeddings are learned along with model tuning. Note that the templates in WARP are also trainable, but here we only use its soft verbalizers. In single verbalizer experiments, we initialize the token embeddings randomly for fairness. And in extra verbalizer experiments, they are initialized with label names.

\begin{table}[h]
    \centering
    \resizebox{\linewidth}{!}{
    \begin{tabular}{c|l|cccc}
         \toprule  
         $K$ & Method & AG & DB & Yahoo & Few\\
         \midrule 
         0 & ManualVerb & \textit{75.13} & \textit{67.06 } & \textit{43.11} & \textit{20.00}\\
         \midrule 
         \multirow{4}{*}{1} & ManualVerb & \textit{76.67} & \textit{85.47} & \textit{50.22} & \textit{41.68} \\
         & SearchVerb & 41.50 & 60.06 & 27.39 & 20.88 \\
         & SoftVerb & 49.79  & 65.35  & 22.72  & 18.78  \\
         \cmidrule(lr){2-6}
         & ProtoVerb & {\bf 64.19}& {\bf 72.85} & {\bf 36.12} & {\bf 25.00}\\
         \midrule 
         \multirow{4}{*}{2} & ManualVerb & \textit{81.06} & \textit{93.61} & \textit{58.65} & \textit{46.44}\\
         & SearchVerb & 65.82  & 78.21 & 40.71 & 31.28 \\
         & SoftVerb & 56.37 & 80.69  & 30.72 & 32.80  \\
         \cmidrule(lr){2-6}
         & ProtoVerb & {\bf 77.34} & {\bf 85.49} & {\bf 46.30} & {\bf 35.72}\\

         \midrule 
         \multirow{4}{*}{4} & ManualVerb & \textit{84.73} & \textit{95.83} & \textit{61.41} & \textit{52.54}\\
         & SearchVerb & 77.43  & 86.40 & 51.58 & 43.10 \\
         & SoftVerb & 74.38 & 89.12 & 41.62 & {\bf 48.77} \\
         \cmidrule(lr){2-6}
         & ProtoVerb & {\bf 81.65} & {\bf 90.91} & {\bf 55.08} &  48.28 \\

         \midrule 
         \multirow{4}{*}{8} & ManualVerb & \textit{85.85} & \textit{96.46} & \textit{64.12} & \textit{56.59}\\
         & SearchVerb & 82.17& 88.41 & 58.64 & 50.78\\
         & SoftVerb & 79.35 & 93.69 & 46.82 & 53.78  \\
         \cmidrule(lr){2-6}
         & ProtoVerb & {\bf 84.03} & {\bf 95.75} & {\bf 61.40} & {\bf 56.06}\\

         \midrule 
         \multirow{4}{*}{16} & ManualVerb & \textit{84.74} & \textit{96.05} & \textit{58.77} & \textit{61.17}\\
         & SearchVerb & 83.40 & 92.00 & 59.66 & 55.49 \\
         & SoftVerb & 80.57  & 86.90  & 58.20  & 58.87 \\
         \cmidrule(lr){2-6}
         & ProtoVerb & {\bf 84.48} & {\bf 96.30} & {\bf 64.35} & {\bf 61.29}\\

         \bottomrule
    \end{tabular}
    }
    
    \caption{Results for single verbalizer experiments. We report the mean accuracy scores (\%) over 3 random seeds. \textit{Italic}: results with task-specific knowledge. \textbf{Bold}: best results without task-specific knowledge.}
    \label{tab:single}
\end{table}

\begin{table}[!ht]
    \centering
    \resizebox{\linewidth}{!}{%
    \begin{tabular}{c|l|cccc}
         \toprule  
         $K$ & Method & AG & DB & Yahoo & Few\\
         \midrule 
         0 & ManualVerb & 75.13 & 67.06 & 43.11 & 20.00 \\
         \midrule 
         \multirow{6}{*}{1} & Fine-tuning &25.45 & 10.80 & 10.59 & 7.48 \\
         & ManualVerb & 76.67 & 85.47 & {\bf 50.22} & 41.68 \\
         & SearchVerb+ & 51.82 & 81.31 & 43.24 & 35.64 \\
         & SoftVerb+ & 76.34 & 85.85 & 49.11 & 37.66 \\
         \cmidrule(lr){2-6}
         & ProtoVerb+ & {\bf 77.71}& {\bf 88.16} & 50.08 & {\bf 43.20}\\
         & \quad w/o tuning & 76.28 & 78.32 & 45.01 & 29.51 \\
         \midrule 
         \multirow{6}{*}{2} & Fine-tuning & 25.78 & 49.01 & 11.26 & 19.03 \\
         & ManualVerb & 81.06 & 93.61 & 58.65 & 46.44 \\
         & SearchVerb+ & 77.56 & 91.79 & 52.46 & 42.13 \\
         & SoftVerb+ & 79.95 & 93.68 & 55.73 & 42.17 \\
         \cmidrule(lr){2-6}
         & ProtoVerb+ & {\bf 84.09} & {\bf 94.77} & {\bf 59.33} & {\bf 48.69}\\
         & \quad w/o tuning & 82.13 & 86.11 & 50.34 & 34.44 \\
         \midrule 
         \multirow{6}{*}{4} & Fine-tuning &28.14 & 94.08 & 26.02 & 20.98 \\
         & ManualVerb & 84.73 & 95.83 & 61.41 & 52.54 \\
         & SearchVerb+ & 81.25 & 95.16 & 58.98 & 50.61 \\
         & SoftVerb+ & 84.22 & 94.90 & 59.01 & 49.45 \\
         \cmidrule(lr){2-6}
         & ProtoVerb+ & {\bf 85.71} & {\bf 96.74} & {\bf 66.14} & {\bf 54.16}\\
         & \quad w/o tuning & 83.05 & 89.56 & 55.59 & 35.55 \\
         \midrule 
         \multirow{6}{*}{8} & Fine-tuning & 72.78 & 96.83 & 54.76 & 49.77\\
         & ManualVerb & 85.85 & 96.46 & 64.12 & 56.59 \\
         & SearchVerb+ & 85.68 & 97.57 & 65.32 & 56.58 \\
         & SoftVerb+ & 86.54 & 97.40 & 63.48 & 54.30 \\
         \cmidrule(lr){2-6}
         & ProtoVerb+ & {\bf 87.25} & {\bf 97.64} & {\bf 66.61} & {\bf 58.30}\\
         & \quad w/o tuning & 83.79 & 92.61& 59.42& 34.37 \\
         \midrule 
         \multirow{6}{*}{16} & Fine-tuning & 84.14& {\bf 97.25} & 64.27 & 52.66\\
         & ManualVerb & 84.74 & 96.05& 58.77 & 61.17\\
         & SearchVerb+ & 85.30 & 95.08 & 59.34 & 61.70 \\
         & SoftVerb+ & 85.65 & 96.34 & 58.68 & 59.23 \\
         \cmidrule(lr){2-6}
         & ProtoVerb+ & {\bf 87.98} & 97.22& {\bf 65.65} & {\bf 62.55}\\
         & \quad w/o tuning & 84.78& 93.46 & 60.89 & 33.96 \\
         \bottomrule
    \end{tabular}%
    }
    
    \caption{Results for multiple verbalizer experiments. We report the mean accuracy scores (\%) over 3 random seeds. ProtoVerb+ w/o tuning: apply ProtoVerb to untuned PLMs. \textbf{Bold}: best results.}
    \label{tab:extra}
\end{table}

\subsection{Single Verbalizer Results}
\label{sec:single}
Table~\ref{tab:single} presents the performance of different verbalizers. Overall, ManualVerb is the most powerful verbalizer, which is reasonable because it is picked by human with domain knowledge.
ProtoVerb outperforms SearchVerb and SoftVerb remarkably and consistently, especially when only 1 or 2 instances per class are given. The poor performances of the two baselines under extreme data scarcity corroborate the issues we claim in~\cref{sec:intro}. As the training data become sufficient, ProtoVerb gets comparable or even exceeding scores compared with ManualVerb, showing that ProtoVerb is able to learn prototypes that well represent the classes. At the same time, the gaps between ManualVerb and other verbalizers narrow, which also indicates that we can summarize data across various ways.

Across tasks, ProtoVerb gets better results on topic classification than entity typing. A possible reason is that FewNERD is a fine-grained entity typing dataset, in which the differences across classes are subtle. For example, it is hard for ProtoVerb to discriminate between ``person-artist/author'' and ``person-director'' with only a few instances. However, ProtoVerb can also catch up with ManualVerb with enough samples.

\subsection{Multiple Verbalizer Results}
\label{sec:extra}
Table~\ref{tab:extra} shows the experiment results when we ensemble manual verbalizers with automatic verbalizers. The ensembled versions are denoted as SearchVerb+, SoftVerb+, and ProtoVerb+ respectively. From the table, we have the following observations:
(1) Basically, prompt-based tuning outperforms fine-tuning by a large margin with few samples (1$\sim$2 per class). When sufficient training data is available, fine-tuning models will produce comparable results.
(2) Overall, ProtoVerb+ certainly improves the performance of prompt-based tuning under most cases, which demonstrates the effectiveness of ProtoVerb+. At the same time, SearchVerb+ and SoftVerb+ seldom show enhancement compared with ManualVerb. As ProtoVerb+ does not introduce any external knowledge, this illustrates that ProtoVerb+ provides a better way to utilize training data. 

Finally, we also present the results of applying ProtoVerb+ on untuned PLMs.
It is worth noting that even for untuned models, ProtoVerb+ also boosts them considerably on all tasks. For example on DBPedia, showing only one instance per class to PLMs with ProtoVerb+ leads to 11.26\% absolute accuracy improvement. On topic classification, when more training samples are given, untuned PLMs achieve competitive scores. This observation indicates a new cost-efficient way to leverage training data, which we highlight as valuable for future study of \textbf{none-tuning} methods for PLMs. Compared to the ``in context learning'' in GPT-3~\citep{Brown20language}, ProtoVerb+ is not limited by input length and can deal with arbitrary number of samples. We further study this ``fixed model'' scenario in~\cref{sec:fix}.

\section{Analysis}
In this section, we discuss several analytical topics for further understandings of ProtoVerb. For simplicity, we conduct experiments on AG's News dataset. 

\subsection{Fixed Model Experiments}
\label{sec:fix}

\begin{figure}[h]
    \centering
    \includegraphics[width=\linewidth]{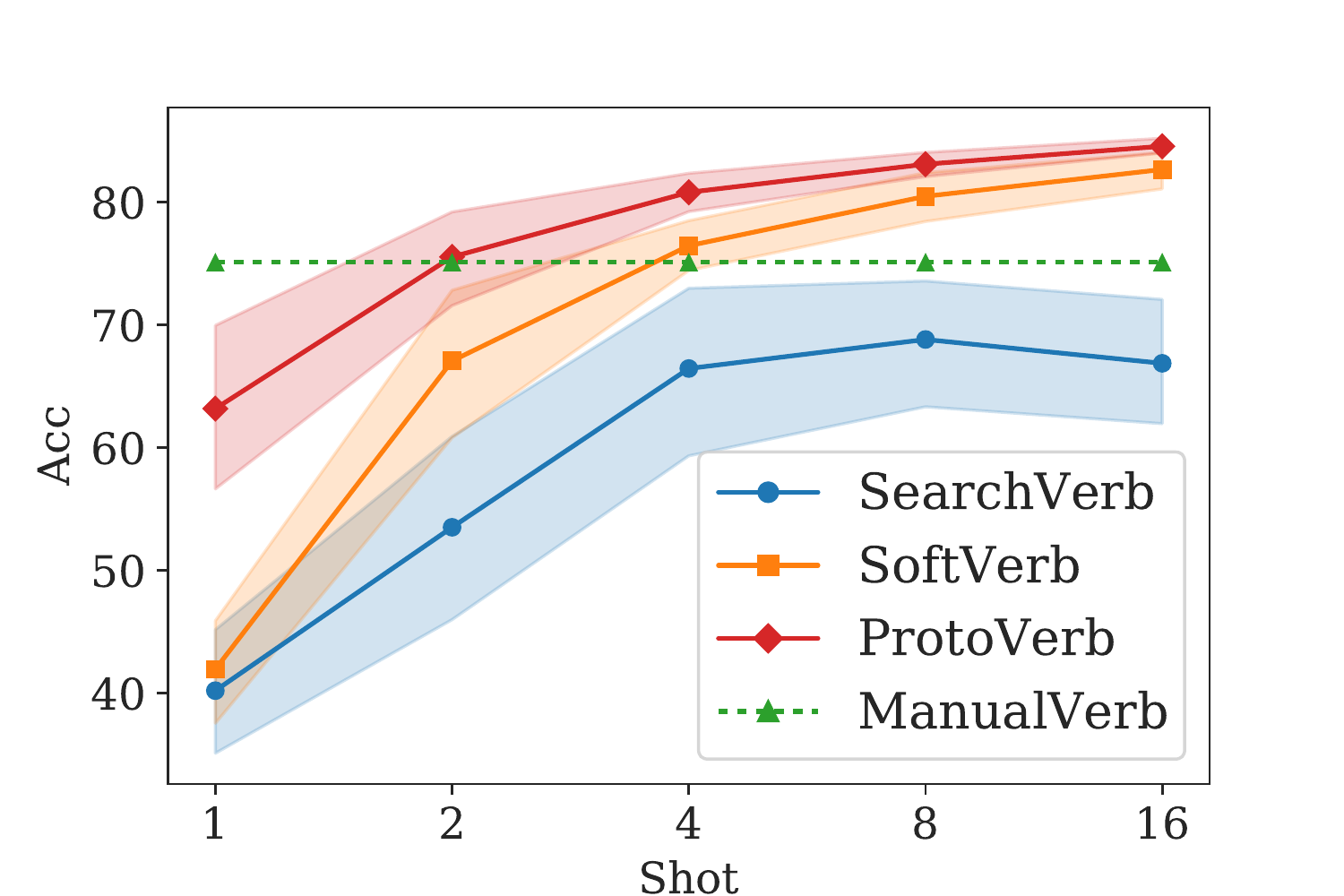}
    \caption{Experiment results with fixed PLMs.  We report the mean accuracy (\%) with 95\% confidence interval on AG's News.}
    \label{fig:untune}
\end{figure}

In \cref{sec:extra}, we see ProtoVerb is still powerful with fixed PLMs. For further comparisons, we conduct experiments to quantitatively evaluate verbalizers when PLMs are fixed. Figure~\ref{fig:untune} gives the results. To clarify, using ManualVerb on fixed PLMs equals the zero-shot setting, which we plot with a dashed line. Meanwhile, different from \cref{sec:extra}, ProtoVerb here is a single verbalizer. From the figure we can conclude that (1) Similar with \cref{sec:single}, ProtoVerb outperforms SoftVerb and SearchVerb by a large margin under low-shot settings. Notably, ProtoVerb exceeds ManualVerb with only 2 shots per class, illustrating the experessive power of prototypes. (2) SoftVerb is also better than SearchVerb under this setting, demonstrating that tunable verbalizers could exploit training data better with PLMs fixed.

\begin{table}[h]
    \centering
    \begin{tabular}{l|ccc}
    \toprule
        Method & $K=2$ & $K=4$ & $K=8$\\
        \midrule
       $\mathcal{L}_{\text{ins}} + \mathcal{L}_{\text{proto}}$  & {\bf 77.34}& {\bf 81.65} & {\bf 84.03} \\
       $\mathcal{L}_{\text{proto}}$ & 76.37& 81.06 & 82.91\\
       Instance Mean  & 73.36 & 77.76 & 82.57 \\
       \bottomrule
    \end{tabular}
    
    \caption{Ablation study of ProtoVerb on AG's News. Instance Mean: using the mean embeddings of instances as prototype embeddings. \textbf{Bold}: best results}
    \label{tab:ablation}
\end{table}

\begin{table}[h]
    \centering
    \begin{tabular}{c|l|ccc}
    \toprule
     \multirow{2}{*}{$K$} & \multirow{2}{*}{ Method} & \multicolumn{3}{|c}{\# Noisy Samples}\\
     \cmidrule{3-5}
     & & 1 & 2 & 3  \\
        \midrule
     \multirow{3}{*}{8} &  SearchVerb & 4.86 & 5.96 & 5.19 \\
      &  SoftVerb  & 4.84 & 7.80 & 11.71 \\
      & ProtoVerb & {\bf2.34} & {\bf3.11} & {\bf4.37} \\
      \midrule
      \multirow{3}{*}{16} &  SearchVerb & 0.80 & 2.93 & 5.18 \\
      &  SoftVerb  & 2.01 & 4.17 & 4.58 \\
      & ProtoVerb & {\bf 0.04} & {\bf 2.13} & {\bf3.16} \\
       \bottomrule
    \end{tabular}
    \caption{Accuracy drop (\%) with noisy samples. Lower is better. \textbf{Bold}: best results.}
    \label{tab:noisy}
\end{table}

\begin{table*}[h]
    \centering
    \begin{tabular}{l|c|c}
    \toprule
       Class & $K=1$ & $K=16$  \\
       \midrule
        World & Qaida, Syria, Iraq, Nusra, TPP & Taliban, Iraq, Afghan, militants, rebellion \\
        Sports & Steelers, Raptors, Knicks, Dodgers & ball, ESPN, baseball, Fifa, Sports \\
        Business & cash, earnings, Securities, NYSE & Dow, dividend, investing, markets \\
        Tech & LTE, Tel, Huawei, Mbps, VPN & Vault, IBM, Qualcomm, Technologies\\
        \bottomrule
    \end{tabular}
    \caption{Words that are most similar with prototypes of each class on AG's News.}
    \label{tab:words}
\end{table*}

\subsection{Ablation Study}
To validate the effect of each part in the loss function, we conduct an ablation study on AG's News dataset. For comparison, we consider two variants of prototype calculation methods: (1) ProtoVerb with $\mathcal{L}_{\text{proto}}$ only. (2) Following ProtoNet~\citep{Snell17proto}, take the average of instance embeddings for prototype embeddings.
Table~\ref{tab:ablation} shows the results. Compared to taking the mean embedding vectors directly, optimizing the embedding vectors of prototypes using our loss functions leads to better performances and stability. Adding $\mathcal{L}_{\text{ins}}$ is also beneficial, meaning that $\mathcal{L}_{\text{ins}}$ helps ProtoVerb in learning instance embeddings.

\subsection{Robustness on Noisy Samples}

Noisy data are commonly seen as threats in real-world datasets for few-shot learning systems. For automatic verbalizers, noisy data are more harmful because of the effect on both the quality of verbalizers and the training process. In this section, we evaluate the robustness of different automatic verbalizers against noisy samples on AG's News. For training stability, we set $K=8,16$. Table~\ref{tab:noisy} presents the accuracy drop when there are 1, 2, or 3 samples having wrong labels. 
It is clearly seen that a limited number of noisy samples will hinder the performance greatly, showing the vulnerability of automatic verbalizers. Meanwhile, we can also find that ProtoVerb is more robust than baseline methods when facing noisy samples.

\subsection{Prototype Discretization}
\label{sec:discrete}
Since ProtoVerb learns continuous prototype vectors, their meanings are implicit. Here we manage to investigate which words are most similar to the learned prototypes. Due to word embeddings and prototype vectors lying in different embedding spaces, we can not directly calculate their similarity. Hence we use the vocabulary as the input texts (one word at a time) to get the top-scored word for each class. On AG's News dataset, we collect some most similar words for each class and list them in Table~\ref{tab:words}. 

To investigate the property of prototypes learned with different numbers of samples, we present words for $K=1$ and $K=16$. With the table, we see that: (1) Even when only one example is available, the learned prototypes are meaningful. Most of the similar words are proper nouns and entity names closely related to class topics. For example, ``Steelers'', ``Raptors'', ``Knicks'', and ``Dodgers'' are all baseball or basketball teams that appear frequently in sports news. We attribute this to prompt mechanism that allows PLMs to extract the most conclusive information and fill the \texttt{[MASK]} with it. Then the relevant words are also included.
(2) With more training instances, prototypes show diverse interests. Despite entity names, more ``conceptual'' words show up on the list, such as ``ball'' and ``Sports'' for class Sports. We interpret this as the summarization and abstraction ability of prototypes. Given many instances, prototypes are enforced to capture their common features, hence some abstract concepts are found automatically. In this way, ProtoVerb encapsulates class-level, rather than entity-level, semantics, which leads to better performance on unseen data.

\subsection{Is ProtoVerb Similar with ManualVerb?}
\begin{figure}[h]
    \centering
    \includegraphics[width=\linewidth]{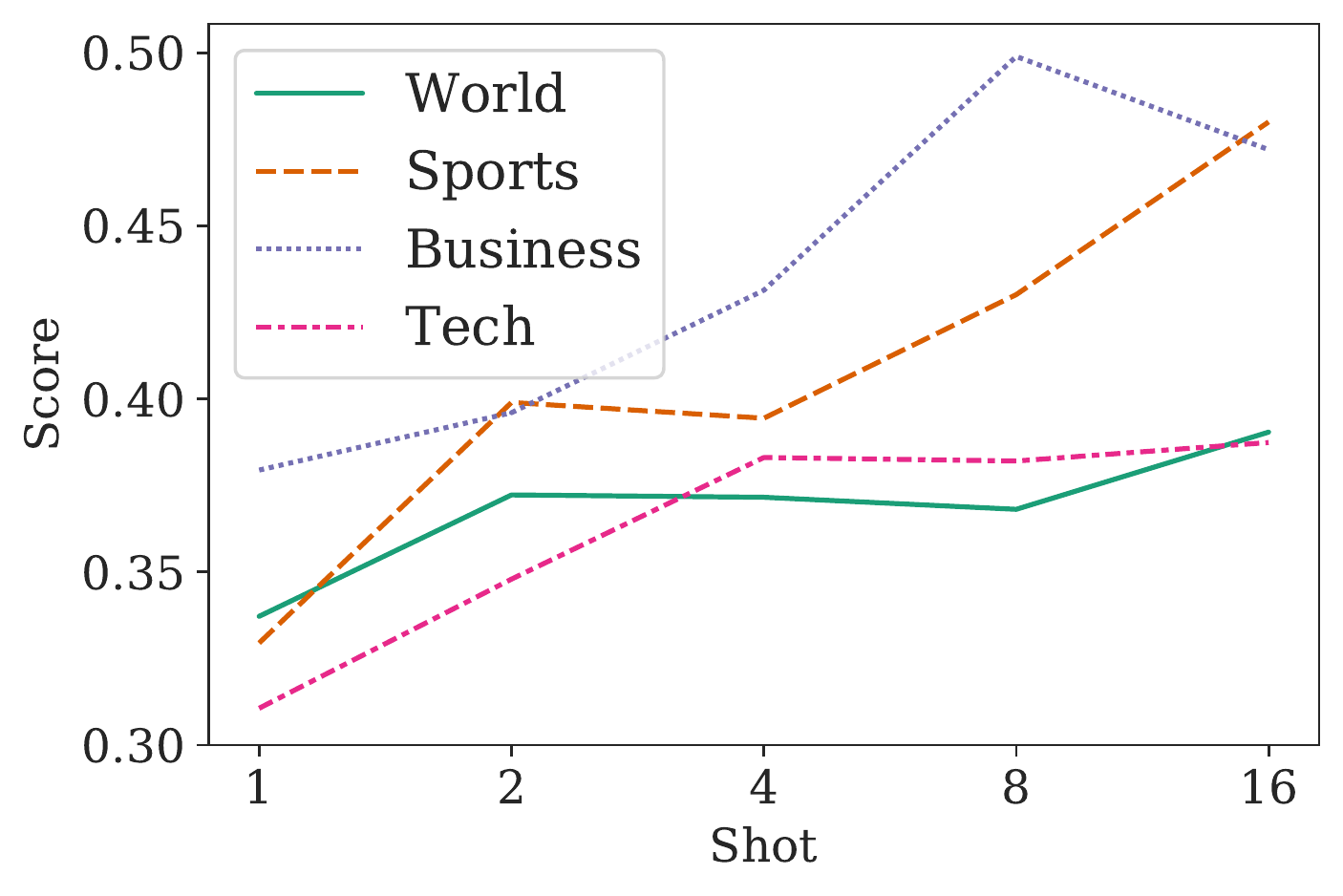}
    \caption{Similarity scores between ProtoVerb and ManualVerb on AG's News.}
    \label{fig:label_vocab}
\end{figure}

To give further analyses for the inner workings of prototypes, we measure the similarity between ProtoVerb and ManualVerb to see whether ProtoVerb is able to learn abstract concepts as humans do. On AG's News dataset, we calculate the similarity scores between prototypes and manual verbalizers and normalize the scores using the softmax function across the four classes. In Figure~\ref{fig:label_vocab} we plot the scores with various shots. It is clearly seen that the similarity of prototypes and corresponding verbalizers are above average (0.25). As shot increases, the scores also gradually grow, which illustrates that prototypes can capture the conceptual information better from more instances. This observation matches our findings in ~\cref{sec:discrete}. Among the four classes, Business and Sports get higher scores than World and Tech. A reasonable guess is that World and Tech news includes diverse sub-topics that are hard to summarize.

\section{Conclusion}
In this paper, we propose a novel approach for automatic verbalizer construction in prompt-based tuning. The proposed ProtoVerb learns class prototypes from training instances using contrastive learning. We explore the performance of ProtoVerb on few-shot topic classification and entity typing tasks. As a single verbalizer, ProtoVerb outperforms state-of-the-art automatic verbalizers considerably. Working together with manual verbalizers, ProtoVerb can also consistently improve prompt-based tuning with minor effort. The results validate the effectiveness of ProtoVerb. Our analysis further reveals the intrinsic properties of prototypes. For future work, we will focus on extending ProtoVerb for effective non-tuning algorithms of PLMs and prompt-tuning with soft templates. Moreover, we are finding proper ways to combine label words and prototypes for verbalizer construction.

\section*{Acknowledgements}
This work is supported by the National Key R\&D Program of China (No. 2020AAA0106502), Institute for Guo Qiang at Tsinghua University, Beijing Academy of Artificial Intelligence (BAAI), and International Innovation Center of Tsinghua University, Shanghai, China. Ganqu Cui and Shengding Hu conducted the experiments. Ganqu Cui, Shengding Hu, Ning Ding and Zhiyuan Liu wrote the paper. Longtao Huang provided valuable advices to the research.

\section*{Ethical Considerations}
Our work explores how to stimulate large PLMs with few samples. We conduct experiments under the few-shot setting, where requires less training time and fewer resources than normal full-data setting. Also, we open up our codes and hyperparameters to facilitate future reproduction without repeated energy cost.

\bibliography{anthology}
\bibliographystyle{acl_natbib}

\appendix

\section{Templates}
\label{sec:appendix}
For topic classification, we use the default templates and verbalizers in OpenPrompt~\cite{ding2021openprompt}.

\textbf{AG's News} is a news' topic classification dataset. There are four categories: World, Sports, Business, and Tech. We use the following templates.
\begin{align*}
\mathcal{T}_1(x)&= \text{A \texttt{[MASK]} news: } x \\
{}\mathcal{T}_2(x)&= x\ \text{This topic is about \texttt{[MASK]}.} \\
{}\mathcal{T}_3(x)&= \text{[ Category : \texttt{[MASK]} ] } x \\
{}\mathcal{T}_4(x)&= \text{[ Topic : \texttt{[MASK]} ] } x 
\end{align*}

\textbf{DBPedia} is an ontology classification dataset. Each sample contains an article title $x$ and abstract $y$ extracted from Wikipedia, and the task is to classify the subject's  ontology class. There are 14 classes in total. We employ four templates shown below:
\begin{align*}
    \mathcal{T}_1(x,y)&= x\ y\ x\ \text{is a \texttt{[MASK]}.} \\
    \mathcal{T}_2(x,y)&= x\ y\ \text{In this sentence,} x \text{is a \texttt{[MASK]}.} \\
    \mathcal{T}_3(x,y)&= x\ y\ \text{The type of} x \text{is \texttt{[MASK]}.} \\
    \mathcal{T}_4(x,y)&= x\ y\ \text{The category of} x \text{is \texttt{[MASK]}.}
\end{align*}

\textbf{Yahoo} is a question classification dataset with 10 classes. Each piece of text consists of a question and an answer. We use the templates in AG's News where ``news'' is replaced with ``question'' in $\mathcal{T}_1(\cdot)$
\begin{align*}
    \mathcal{T}_1(x)&= \text{A \texttt{[MASK]} question: } x \\
    \mathcal{T}_2(x)&= x\ \text{This topic is about \texttt{[MASK]}.} \\
    \mathcal{T}_3(x)&= \text{[ Category : \texttt{[MASK]} ] } x \\
    \mathcal{T}_4(x)&= \text{[ Topic : \texttt{[MASK]} ] } x
\end{align*}

\textbf{FewNERD} is a large-scale fine-grained entity typing dataset with 66 types and we use the official split of its supervised setting. Following~\citep{ding2021prompt}, we employ 3 templates as below
\begin{align*}
    \mathcal{T}_1(x)&= x\ \text{\texttt{[ENT]} is \texttt{[MASK]}.} \\
    \mathcal{T}_2(x)&= x\ \text{\texttt{[ENT]} is a \texttt{[MASK]}.} \\
    \mathcal{T}_3(x)&= x\ \text{In this sentence, \texttt{[ENT]} is a \texttt{[MASK]}.}
\end{align*}
where \texttt{[ENT]} copies the entity mention in the sentence.

\end{document}